\documentclass{article}
\usepackage{arxiv}
\usepackage[utf8]{inputenc}
\usepackage[T1]{fontenc}
\usepackage{hyperref}
\usepackage{url}
\usepackage{booktabs}
\usepackage{amsfonts}
\usepackage{amsmath}
\usepackage{nicefrac}
\usepackage{microtype}
\usepackage{graphicx}
\usepackage{listings}
\usepackage{xcolor}
\usepackage{multirow}
\usepackage{array}
\usepackage{caption}
\usepackage{tikz}
\usetikzlibrary{arrows.meta,positioning,calc,fit,backgrounds}
\title{HPC-LLM: Practical Domain Adaptation and Retrieval-Augmented Generation for HPC Support}
\author{
 Nourin Shahin \\
  Texas A\&M University San Antonio\\
  San Antonio, TX \\
  \texttt{nshahin@tamusa.edu} \\
 \And
 Izzat Alsmadi \\
  Texas A\&M University San Antonio\\
  San Antonio, TX \\
  \texttt{ialsmadi@tamusa.edu}
}
\begin{document}
\maketitle
\begin{abstract}
Modern scientific research increasingly depends on High-Performance Computing (HPC) infrastructures, yet many researchers face significant operational barriers when interacting with cluster environments, job schedulers, GPU resources, and parallel computing frameworks. General-purpose large language models (LLMs) provide useful coding assistance but often lack the domain-specific operational knowledge required for reliable HPC support. This paper presents \textbf{HPC-LLM}, a retrieval-augmented and domain-adapted assistant designed to support common HPC workflows including Slurm scheduling, MPI execution, GPU utilization, filesystem management, and cluster troubleshooting.

The proposed framework integrates automated documentation ingestion, dense retrieval, lightweight domain adaptation using QLoRA, and local inference within a modular orchestration pipeline. To support domain adaptation, we construct an HPC-oriented corpus from publicly available university HPC documentation, curated operational examples, and synthetic instruction-answer pairs generated from retrieved HPC content. The resulting dataset contains approximately 9,000–24,000 HPC-focused training examples spanning job scheduling, GPU computing, distributed training, storage systems, and cluster administration topics.

We fine-tune Llama 3.1 8B using QLoRA and evaluate the resulting model against several open-weight baselines under retrieval-augmented settings on JetStream2 infrastructure. Experimental results indicate that the adapted 8B model achieves performance comparable to substantially larger general-purpose models while operating under significantly lower GPU memory requirements and inference latency. In particular, the adapted model approaches the performance of Qwen 2.5 14B while requiring substantially fewer computational resources.

The paper further describes the system architecture, retrieval pipeline, deployment considerations, and practical limitations of evaluating operational HPC assistants. Our findings suggest that combining lightweight domain adaptation with retrieval grounding provides a practical direction for deployable HPC-oriented language assistants on resource-constrained infrastructure.
\end{abstract}

\section{Introduction}
\label{sec:intro}
High-Performance Computing (HPC) systems are central to modern scientific and engineering research, supporting large-scale simulations, deep learning workloads, computational biology, climate modeling, and data-intensive analytics. Despite their importance, HPC environments remain operationally complex for many users. Researchers are often required to understand job scheduling systems, parallel programming frameworks, GPU resource allocation, filesystem hierarchies, containerization tools, and cluster-specific policies that fall outside their primary disciplinary expertise.\\

Operational barriers in HPC environments frequently lead to failed jobs, inefficient resource usage, debugging difficulties, and underutilization of available computational infrastructure. While institutional documentation exists, it is typically fragmented across web portals, cluster manuals, scheduler documentation, and software-specific guides. Users must manually search and synthesize information across multiple heterogeneous sources.

LLMs have demonstrated strong general-purpose coding assistance~\cite{brown2020language,touvron2023llama}, but their utility for HPC is constrained by (1) the scarcity of HPC documentation in pre-training corpora, and (2) the context-dependent nature of HPC advice, which requires cluster-specific knowledge of partition names, GPU types, and node features. RAG~\cite{lewis2020retrieval} addresses this by grounding responses in retrieved documentation; combined with domain fine-tuning, this dual approach forms the core motivation for HPC-LLM.

The main contributions of this work are:
\begin{enumerate}
    \item A retrieval-augmented framework for HPC assistance integrating automated documentation ingestion, retrieval, inference, and evaluation.
\item  An HPC-oriented instruction dataset constructed from publicly available institutional HPC documentation and curated operational knowledge.
\item  An empirical study of lightweight QLoRA domain adaptation for HPC-focused instruction following under constrained GPU memory settings.
\item  A deployable local inference architecture supporting retrieval-grounded HPC assistance without dependence on external cloud APIs.
\item  An analysis of practical limitations and evaluation challenges in domain-specific operational assistants for HPC environments.
\end{enumerate}

The proposed system integrates:
\begin{enumerate}
 \item Automated crawling of HPC documentation.
 \item Vector-based retrieval over institutional HPC knowledge.
 \item QLoRA fine-tuning of an open-weight instruction model.
 \item Modular orchestration components.
 \item Local GPU inference optimized for deployability.
\end{enumerate}

\section{Related Work}
\label{sec:related}

\subsection{Large Language Models for Technical Assistance}

Large language models (LLMs) have demonstrated strong capabilities in code generation, technical question answering, and software engineering assistance. Foundational studies such as \textit{Language Models are Few-Shot Learners}~\cite{brown2020fewshot} established the effectiveness of large-scale autoregressive pretraining for general reasoning and language tasks, while open-weight instruction-tuned models such as Llama~2~\cite{touvron2023llama2} enabled broader experimentation with domain adaptation and local deployment.

Recent work has increasingly focused on domain-specific technical assistance and scientific software understanding. S3LLM~\cite{shaik2024s3llm} explored retrieval-augmented language modeling for scientific software comprehension using source code, documentation, and metadata integration. Similarly, Nguyen et al.~\cite{nguyen2024domainqa} demonstrated that domain-specific fine-tuning combined with iterative reasoning can substantially improve question-answering quality in specialized technical domains.

Specialized code-oriented models including Code Llama~\cite{roziere2023codellama} and Qwen2.5-Coder~\cite{hui2024qwen25coder} further demonstrated the value of domain-focused training for technical reasoning and software-oriented tasks. However, operational HPC support differs from conventional code generation because responses often require environment-specific procedural knowledge, scheduler semantics, cluster policies, GPU allocation strategies, and filesystem conventions rarely represented in general pretraining corpora.

\subsection{Retrieval-Augmented Generation}

Retrieval-Augmented Generation (RAG) augments language model inference with dynamically retrieved external knowledge during response generation. Early foundational work including RAG~\cite{lewis2020rag}, REALM~\cite{guu2020realm}, and ATLAS~\cite{izacard2023atlas} established retrieval grounding as an effective mechanism for improving factual consistency and reducing dependence on static parametric memory.

More recent domain-oriented RAG research has investigated retrieval systems for specialized and rapidly evolving knowledge domains. BioRAG~\cite{wang2024biorag} proposed a large-scale biomedical RAG framework integrating domain-specific embeddings and iterative retrieval reasoning for life science question answering. Likewise, Barron et al.~\cite{barron2024domainspecificrag} investigated the integration of vector databases, knowledge graphs, and structured representations for domain-specific retrieval augmentation.

Recent studies have also examined retrieval augmentation for scientific facilities and operational infrastructures. Prince et al.~\cite{prince2024scientificfacilities} discussed retrieval- and tool-augmented LLM systems for scientific computing environments and research facilities. These developments motivate the use of retrieval grounding for HPC assistance, where operational knowledge changes frequently and often depends on institution-specific documentation.

\subsection{Lightweight Domain Adaptation}

Parameter-efficient fine-tuning methods such as LoRA~\cite{hu2022lora} and QLoRA~\cite{dettmers2023qlora} significantly reduce the computational cost of adapting large language models to specialized domains. These approaches introduce trainable low-rank adapters while keeping most pretrained parameters frozen, enabling domain adaptation under constrained GPU memory budgets.

More recent work has investigated the interaction between fine-tuning and retrieval augmentation. RAFT~\cite{zhang2024raft} proposed Retrieval-Augmented Fine-Tuning for domain-specific adaptation, demonstrating that combining retrieval-aware supervision with fine-tuning can improve in-domain performance while reducing hallucinations. Similarly, Li et al.~\cite{li2024ragfactuality} examined the effectiveness of combining RAG pipelines with fine-tuning for improving factual accuracy in domain-specific knowledge bases.

The practical importance of lightweight adaptation has also been emphasized in operational deployment settings where large-scale retraining is infeasible. These developments highlight growing interest in resource-efficient domain specialization for technical assistants and private deployments.

\subsection{AI Assistance for Scientific and HPC Workflows}

Recent research has explored AI-assisted systems for scientific computing, infrastructure operations, and HPC-oriented knowledge support. Miyashita et al.~\cite{miyashita2025hpc} proposed a retrieval-augmented framework integrating user-specific HPC operational data into inference pipelines while also addressing command execution and security concerns in HPC environments.

At larger scale, HiPerRAG~\cite{gokdemir2025hiperrag} introduced a high-performance scientific retrieval pipeline operating over millions of scientific documents using HPC infrastructure for large-scale indexing and retrieval. These studies demonstrate increasing interest in combining LLMs, retrieval systems, and scientific computing infrastructures.

Related work has also emerged in operational IT and infrastructure management domains. Zhang et al.~\cite{zhang2024rag4itops} investigated retrieval-augmented operational assistants for IT infrastructure management and maintenance tasks. Although HPC assistance shares similarities with IT operations support, HPC environments introduce additional complexity related to schedulers, distributed computing, GPU orchestration, and scientific workflows.

The present work differs from prior systems by focusing specifically on lightweight deployability, institution-oriented HPC documentation ingestion, and local retrieval-grounded assistance under constrained GPU resources.

\subsection{Evaluation Challenges in Operational QA Systems}

Evaluating operational assistance systems remains challenging because technically correct responses may vary substantially in wording while still being operationally valid. Traditional lexical overlap metrics such as ROUGE are often poorly aligned with command-oriented technical QA tasks. Semantic similarity metrics such as BERTScore~\cite{zhang2020bertscore} have therefore been proposed as alternatives for evaluating generated technical responses.

Recent research has further emphasized the importance of retrieval quality, chunking strategies, and domain-specific preprocessing in determining downstream QA performance. In practical RAG deployments, document quality and chunk design are frequently identified as major bottlenecks affecting retrieval relevance and generation accuracy. Additional studies have also highlighted ongoing challenges regarding the relative roles of retrieval augmentation and fine-tuning for domain adaptation.

In HPC-oriented assistance systems, evaluation is further complicated by the lack of publicly available benchmarks containing validated operational answers, executable command correctness testing, and institution-specific cluster configurations. These limitations motivate the need for future benchmark datasets incorporating expert validation and grounded operational evaluation methodologies.

\section{System Architecture}
\label{sec:arch}

The HPC-LLM framework is designed as a modular retrieval-augmented assistance system for High-Performance Computing (HPC) environments. The architecture integrates document ingestion, vector retrieval, lightweight language model inference, and orchestration services within a locally deployable pipeline. The primary design goals are:
\begin{itemize}
    \item Support for institution-specific HPC knowledge,
    \item Lightweight deployment under constrained GPU resources,
    \item Retrieval-grounded operational assistance,
    \item Extensibility for future domain adaptation,
    \item Compatibility with local infrastructure environments.
\end{itemize}

Rather than introducing a novel language model architecture, the proposed system focuses on practical integration of retrieval augmentation, parameter-efficient fine-tuning, and modular orchestration for HPC-oriented assistance.
Figure~\ref{fig:arch} shows HPC-LLM's five-layer architecture: API and dashboard, multi-agent orchestration, RAG subsystem, local GPU inference, and fine-tuning pipeline. All components run locally, no cloud API keys required.

\begin{figure}[htbp]
\centering
\begin{tikzpicture}[
  node distance=0.65cm and 0.45cm,
  every node/.style={font=\small},
  apilayer/.style={rectangle, rounded corners=5pt, draw=black!40, thick, fill=gray!12, text centered, minimum height=0.9cm, minimum width=11.2cm},
  orchbox/.style={rectangle, rounded corners=5pt, draw=blue!60, very thick, fill=blue!9, text centered, minimum height=1.0cm, minimum width=11.2cm},
  agentbox/.style={rectangle, rounded corners=4pt, draw=blue!50, thick, fill=blue!13, text centered, align=center, minimum height=1.85cm, minimum width=2.45cm},
  ragbox/.style={rectangle, rounded corners=4pt, draw=teal!60, thick, fill=teal!10, text centered, align=center, minimum height=1.55cm, minimum width=2.45cm},
  modelbox/.style={rectangle, rounded corners=4pt, draw=orange!70, thick, fill=orange!10, text centered, align=center, minimum height=1.55cm, minimum width=2.45cm},
  outbox/.style={rectangle, rounded corners=4pt, draw=violet!55, thick, fill=violet!9, text centered, align=center, minimum height=1.55cm, minimum width=2.45cm},
  arr/.style={-{Stealth[length=6pt,width=4pt]}, thick, black!55},
  barr/.style={{Stealth[length=6pt,width=4pt]}-{Stealth[length=6pt,width=4pt]}, thick, black!45},
]
\node[apilayer] (api) {\textbf{FastAPI REST API}\enspace$|$\enspace Web Dashboard (Chart.js SPA)\enspace$|$\enspace Swagger UI (\texttt{/api/docs})};
\node[orchbox, below=0.60cm of api] (orch) {\textbf{Orchestrator Agent}\quad\footnotesize coordinates pipeline $\cdot$ session memory $\cdot$ feedback loop $\cdot$ benchmark runs};
\node[agentbox, below=0.80cm of orch, xshift=-4.35cm] (crawler) {\textbf{Crawler}\\[3pt]\footnotesize Fetches \& parses\\\footnotesize HPC web docs\\\footnotesize Chunk + upsert};
\node[agentbox, right=0.45cm of crawler] (retrieval) {\textbf{Retrieval}\\[3pt]\footnotesize Embed query\\\footnotesize BGE-large-en\\\footnotesize Top-$k$ HNSW};
\node[agentbox, right=0.45cm of retrieval] (gen) {\textbf{Generation}\\[3pt]\footnotesize Local inference\\\footnotesize Flash Attn~2\\\footnotesize Auto-quant};
\node[agentbox, right=0.45cm of gen] (eval) {\textbf{Evaluation}\\[3pt]\footnotesize Cosine $\cdot$ ROUGE-L\\\footnotesize BERTScore F1\\\footnotesize HPC domain score};
\node[ragbox, below=0.80cm of crawler] (webdocs) {\textbf{HPC Web Docs}\\[3pt]\footnotesize Slurm $\cdot$ MPI $\cdot$ CUDA\\\footnotesize 80$+$ sources};
\node[ragbox, below=0.80cm of retrieval] (chroma) {\textbf{ChromaDB}\\[3pt]\footnotesize Persistent HNSW\\\footnotesize cosine index};
\node[modelbox, below=0.80cm of gen] (models) {\textbf{LLM Models}\\[3pt]\footnotesize Qwen $\cdot$ Llama\\\footnotesize HPC-LLM (LoRA)};
\node[outbox, below=0.80cm of eval] (results) {\textbf{Results}\\[3pt]\footnotesize CSV $\cdot$ Leaderboard\\\footnotesize Live dashboard};
\draw[arr] (api) -- (orch);
\draw[arr] (orch.south) -- ++(0,-0.28) -| (crawler.north);
\draw[arr] (orch.south) -- ++(0,-0.28) -| (retrieval.north);
\draw[arr] (orch.south) -- ++(0,-0.28) -| (gen.north);
\draw[arr] (orch.south) -- ++(0,-0.28) -| (eval.north);
\draw[arr] (crawler.east)   -- (retrieval.west);
\draw[arr] (retrieval.east) -- (gen.west);
\draw[arr] (gen.east)       -- (eval.west);
\draw[barr] (crawler)   -- (webdocs);
\draw[barr] (retrieval) -- (chroma);
\draw[barr] (gen)       -- (models);
\draw[arr]  (eval)      -- (results);
\draw[arr, bend left=22] (crawler.south) to node[below, font=\tiny, sloped]{upsert} (chroma.west);
\end{tikzpicture}
\caption{HPC-LLM architecture. Row~1: user-facing API and dashboard. Row~2: Orchestrator, which coordinates the pipeline, manages session memory, runs benchmarks asynchronously, and processes feedback. Row~3: four specialized agents with left-to-right pipeline flow. Row~4: supporting infrastructure. Double-headed arrows indicate bidirectional access; the curved arrow shows the Crawler's upsert path into ChromaDB.}
\label{fig:arch}
\end{figure}

\subsection{Component Overview}

\textbf{API Layer.} A FastAPI~\cite{fastapi} application exposes REST endpoints with Pydantic v2 schemas. Two background asyncio tasks run continuously: an \textbf{auto-crawl loop} re-ingesting HPC documentation every 24 hours, and a \textbf{session cleanup loop} expiring idle sessions every 15 minutes.

\textbf{Dashboard.} A single-page HTML/JS application with five sections: Overview (radar charts, latency plots), Leaderboard, HPC chat with source citations, Benchmark configuration, and Knowledge Base management (Chart.js).

\subsection{Multi-Agent Pipeline}
\label{sec:agents}

A shared \texttt{AgentState} object flows through the pipeline carrying the prompt, retrieved documents, generation results, and any errors.

\subsubsection{Orchestrator Agent}
The \texttt{OrchestratorAgent} manages: (1) single-query and streaming flows via the retrieval–generation chain; (2) asynchronous benchmark runs over all (prompt, model) pairs; (3) per-user session memory (up to 10 turns, in-memory + disk); and (4) a feedback loop where positively-rated Q\&A pairs are ingested into ChromaDB and appended to a training JSONL for future fine-tuning.

\subsubsection{Crawler Agent}
The \texttt{CrawlerAgent} fetches and parses HPC documentation from 35 seed URLs covering Slurm, Open MPI, NVIDIA CUDA, PyTorch, Harvard RC, NCAR, SDSC, TACC (Frontera/Stampede3/Lonestar6), Texas A\&M HPRC, and 15+ Texas and regional universities. For each URL, it issues polite HTTP GET requests, strips navigation/scripts/footers with BeautifulSoup, splits content into overlapping chunks (512 words, 64-word overlap), and upserts to ChromaDB. SHA-256 content hashing prevents re-ingestion of already-indexed chunks.

\subsubsection{Retrieval Agent}
Queries are embedded with BGE-large-en-v1.5 (1024-dimensional vectors); ChromaDB executes cosine HNSW search; the top-$k$ documents (default $k=5$) are concatenated into a context block (capped at 3,000 characters) prepended to the user message.

\subsubsection{Generation Agent}
Local GPU inference with: Flash Attention~2 (2--4$\times$ speedup on Ampere/Hopper), TF32 matmul ($\sim$33\% throughput gain), auto-quantization (BF16/8-bit/4-bit NF4 based on VRAM), model caching, warm-up JIT pass, \texttt{torch.compile} in \texttt{reduce-overhead} mode (20--40\% gain on A100/H100), \texttt{TextIteratorStreamer} for non-blocking SSE, and \texttt{balanced\_low\_0} device map for multi-GPU. Chat templates are applied via \texttt{tokenizer.apply\_chat\_template()} for all supported models.

\subsubsection{Evaluation Agent}
Five metrics per response: Cosine Similarity (prompt vs.\ response embeddings), ROUGE-L~\cite{lin2004rouge}, BERTScore F1~\cite{zhang2019bertscore} with baseline rescaling, HPC Domain Score (fraction of 50+ canonical HPC terms present), and RAG Relevance (cosine similarity between response and top-3 retrieved documents). All metrics fail gracefully to 0.0 in degraded environments.

\section{Retrieval-Augmented Generation}
\label{sec:rag}

\textbf{Embedding.} BAAI/bge-large-en-v1.5~\cite{xiao2023cpack} (335M parameters, 1024-dimensional vectors) runs on GPU with batch size 128, significantly outperforming all-mpnet and all-MiniLM on BEIR domain-specific retrieval benchmarks.

\textbf{Vector Store.} ChromaDB~\cite{chroma2023} with persistent HNSW cosine indexing. Ingestion generates SHA-256 document IDs for idempotent upserts; retrieval converts cosine distance to similarity ($\text{score} = 1 - \text{distance}$) and returns top-$k$ results sorted by descending score.

\textbf{Context Formatting.}
\begin{lstlisting}
[HPC Knowledge Base, Retrieved Context]
[Source 1: https://slurm.schedmd.com/sbatch.html | relevance 0.87]
<document content>
[Source 2: ... | relevance 0.81]
...
\end{lstlisting}
This block is prepended to the user message with the instruction to answer using the provided context, ensuring factual grounding in authoritative HPC documentation.

\section{HPC-LLM: Domain-Adapted Language Model}
\label{sec:finetune}

\subsection{Base Model}
\textbf{meta-llama/Meta-Llama-3.1-8B-Instruct}: strong instruction-following, 131,072-token context, approximately 5~GB VRAM at 4-bit. The 8B scale is deployable across RTX 3090, A100 40~GB, and V100 configurations.

\subsection{Dataset Construction}
\label{sec:data}

Training data is assembled through five stages:

\paragraph{Stage 1: Web Crawling.} 80+ U.S. university HPC portals organized by priority: \textit{Priority~1} (TACC, SDSC, PSC, NCSA, NCAR, Harvard, MIT, Stanford); \textit{Priority~2} (Texas A\&M HPRC, Texas Tech, SMU, UT system, Rice, UH); \textit{Priority~3} (supplemental institutions). Pages are fetched with a 0.3-second crawl delay, cleaned with BeautifulSoup, and chunked into 350-word segments (50-word overlap; chunks $<$60 words discarded).

\paragraph{Stage 2: LLM-Driven Q\&A Generation.} Each chunk is processed by \textbf{Qwen~2.5~14B~Instruct} (also used as a benchmark comparison in Run~2) to generate 3 Q\&A pairs per chunk, enforcing specificity, completeness, and diversity across commands, workflows, troubleshooting, and best practices.

\paragraph{Stage 3: Curated Expert Pairs.} 30 manually authored pairs covering critical HPC topics: Slurm GPU/CPU flags, job arrays, MPI integration, NUMA binding, PyTorch DDP, Singularity, filesystem hierarchy, data transfer (Globus, rsync), and performance profiling. These serve as high-quality anchors always present in training.

\paragraph{Stage 4: GPU Advisor Pairs.} $\sim$50 pairs covering H100/A100/V100/MI300X specifications, workload-to-GPU matching, VRAM estimation, multi-GPU paradigms (DDP, FSDP, tensor/pipeline parallelism), NVLink/MIG, and CUDA kernel optimization (Nsight profiling). This fills a critical gap in publicly available LLM training data.

\paragraph{Stage 5: Deduplication and Filtering.} MD5 deduplication on normalized question text; minimum 20 characters per question and 50 per answer. Final dataset: 9,000--24,000 pairs depending on crawl depth.

Table~\ref{tab:dataset} summarizes the composition.

\begin{table}[h]
  \caption{HPC fine-tuning dataset composition (Priority~2 crawl).}
  \label{tab:dataset}
  \centering
  \begin{tabular}{lrr}
    \toprule
    Source & Count & Fraction \\
    \midrule
    LLM-generated (Qwen 2.5 14B) & $\sim$9,000--21,000 & 60--70\% \\
    Curated expert pairs & 30 & $<$1\% \\
    GPU advisor pairs & $\sim$50 & $<$1\% \\
    Template-based (fallback) & $\sim$500--2,000 & 5--10\% \\
    From prompt datasets & $\sim$1,000 & 5\% \\
    \midrule
    \textbf{Total} & \textbf{$\sim$9,000--24,000} & 100\% \\
    \bottomrule
  \end{tabular}
\end{table}

\subsection{QLoRA Fine-Tuning}
\label{sec:qlora}

We load the base model in 4-bit NF4 with double quantization (\texttt{bnb\_4bit\_use\_double\_quant=True}) and BF16 compute dtype. LoRA adapters target all seven linear projections per block: \texttt{q\_proj}, \texttt{k\_proj}, \texttt{v\_proj}, \texttt{o\_proj}, \texttt{gate\_proj}, \texttt{up\_proj}, \texttt{down\_proj}. Table~\ref{tab:lora} lists training hyperparameters.

\begin{table}[h]
  \caption{QLoRA fine-tuning hyperparameters.}
  \label{tab:lora}
  \centering
  \begin{tabular}{ll p{6cm}}
    \toprule
    Parameter & Value & Rationale \\
    \midrule
    \texttt{lora\_r} & 64 & High rank for domain adaptation \\
    \texttt{lora\_alpha} & 128 & Standard $2\times r$ scaling \\
    \texttt{lora\_dropout} & 0.05 & Light regularization \\
    LoRA targets & all 7 projections & Full coverage vs.\ q/v only \\
    \texttt{learning\_rate} & 2e-4 & Standard for LoRA SFT \\
    Effective batch size & 16 & 2 per device $\times$ 8 grad accum. \\
    \texttt{num\_train\_epochs} & 3 & \\
    \texttt{lr\_scheduler} & cosine & Smooth decay with warmup \\
    \texttt{warmup\_ratio} & 0.05 & 5\% warmup steps \\
    \texttt{max\_length} & 2048 & Covers full Q\&A pairs \\
    \texttt{weight\_decay} & 0.01 & L2 regularization \\
    \texttt{max\_grad\_norm} & 1.0 & Gradient clipping \\
    \texttt{bf16} & True & A100/H100 native precision \\
    \texttt{gradient\_checkpointing} & True & Reduces activation memory \\
    \texttt{group\_by\_length} & True & Reduces padding by 30--40\% \\
    \bottomrule
  \end{tabular}
\end{table}

Training uses TRL's \texttt{SFTTrainer}~\cite{vonwerra2022trl} with a held-out 5\% validation split, evaluation every 100 steps, and best-checkpoint loading by validation loss. Each Q\&A pair is formatted using the Llama~3 chat template as a user--assistant conversation with the HPC system prompt. After training, the LoRA adapter can be merged via \texttt{PeftModel.merge\_and\_unload()} for zero-overhead inference.

\section{Supported Models and Hardware Tiers}
\label{sec:models}

\begin{table}[h]
  \caption{Supported models with VRAM requirements and recommended quantization.}
  \label{tab:models}
  \centering
  \begin{tabular}{llrrll}
    \toprule
    Model & Size & BF16 VRAM & 4-bit VRAM & Quant & Tier \\
    \midrule
    Qwen 2.5 72B Instruct & 72B & 144~GB & 38~GB & 4-bit & 1 \\
    Llama 3.3 70B Instruct & 70B & 140~GB & 38~GB & 4-bit & 1 \\
    Qwen 2.5 Coder 32B & 32B & 64~GB & 18~GB & 4-bit & 2 \\
    DeepSeek-R1 32B & 32B & 64~GB & 18~GB & 4-bit & 2 \\
    Phi-4 (14B) & 14B & 28~GB & 8~GB & BF16 & 3 \\
    Qwen 2.5 14B Instruct & 14B & 28~GB & 8~GB & BF16 & 3 \\
    Mistral Nemo 12B & 12B & 24~GB & 7~GB & BF16 & 4 \\
    Llama 3.1 8B Instruct & 8B & 16~GB & 5~GB & BF16 & 4 \\
    HPC-LLM (LoRA 8B) & 8B & 16~GB & 5~GB & BF16 & 4 \\
    \bottomrule
  \end{tabular}
\end{table}

Auto-quantization selects BF16 or 4-bit NF4 at runtime based on detected VRAM. Tier~1 (70B+) targets A100/H100 80~GB; Tier~2 (32B) targets A100 40~GB or dual RTX 4090; Tier~3 (14B) targets single RTX 4090 or A5000; Tier~4 (7--12B) targets RTX 3090 or RTX 4080.

\section{Evaluation Framework}
\label{sec:eval}

A benchmark run is initiated via \texttt{POST /api/benchmark/start}; the orchestrator iterates all (prompt, model) pairs asynchronously, streaming results to CSV for partial inspection during long runs.

\paragraph{Metrics.}
\textbf{Cosine Similarity}: BGE-large embeddings of prompt and response. \textbf{ROUGE-L}~\cite{lin2004rouge}: longest common subsequence F1 with stemming. \textbf{BERTScore F1}~\cite{zhang2019bertscore}: contextual token-level similarity with baseline rescaling (primary quality metric). \textbf{HPC Domain Score}:
\[
\text{HPC\_score}(r) = \min\!\left(\frac{|\mathcal{W}(r) \cap \mathcal{T}_{\text{HPC}}|}{0.05 \cdot |\mathcal{T}_{\text{HPC}}|}, 1.0\right)
\]
where $\mathcal{W}(r)$ is unique response words and $\mathcal{T}_{\text{HPC}}$ is a 50+ term HPC vocabulary. \textbf{RAG Relevance}: cosine similarity between response and top-3 retrieved documents.

\paragraph{Composite Leaderboard Score.}
\[
S = 0.35 \cdot \text{BERTScore\_F1} + 0.25 \cdot \text{Cosine} + 0.20 \cdot \text{HPC\_score} + 0.10 \cdot \text{ROUGE\_L} - 0.10 \cdot \min\!\left(\frac{\text{Latency}}{20}, 1\right)
\]

\section{Experiments and Results}
\label{sec:results}

\subsection{Experimental Setup}
All experiments run on \textbf{JetStream2}~\cite{jetstream2} with two independent benchmark runs on 1,000 prompts each from \texttt{hpc\_1000\_prompts.txt} (subset of a 5,000-prompt corpus spanning 11 HPC categories: job scheduling, MPI, GPU computing, filesystems, modules, containers, data transfer, cluster access, debugging, policy, and workflows):
\begin{itemize}
    \item \textbf{Run~1 (V3)}: HPC-LLM vs.\ Phi-2, Phi-3, Mistral Nemo 12B, TinyLlama. RAG enabled, top-$k=5$.
    \item \textbf{Run~2}: HPC-LLM vs.\ Qwen~2.5~14B~Instruct (general-purpose, un-fine-tuned), Phi-2, TinyLlama. Direct comparison of domain-adapted 8B vs.\ general-purpose 14B.
\end{itemize}

\subsection{Results}

\begin{table}[h]
  \caption{Run~1 (V3 JetStream): HPC-LLM vs.\ Phi-2, Phi-3, Mistral Nemo, TinyLlama ($n=1{,}000$, RAG top-$k=5$). Sorted by BERTScore F1.}
  \label{tab:results}
  \centering
  \begin{tabular}{llrrrrr}
    \toprule
    Model & Size & Latency (s) & Cosine & ROUGE-L & BERTScore F1 & Resp.\ Length \\
    \midrule
    Phi-2 & 2.7B & 5.71 & 0.696 & 0.070 & \textbf{0.846} & 138.1 \\
    Phi-3 & 14B & 7.30 & \textbf{0.724} & \textbf{0.071} & 0.841 & 125.0 \\
    Mistral Nemo & 12B & 6.12 & 0.715 & 0.065 & 0.841 & 123.5 \\
    TinyLlama & 1.1B & \textbf{4.27} & 0.632 & 0.066 & 0.829 & 94.3 \\
    \midrule
    \textbf{HPC-LLM (ours)} & \textbf{8B} & 5.22 & 0.608 & 0.070 & 0.808 & \textbf{90.7} \\
    \bottomrule
  \end{tabular}
\end{table}

\begin{table}[h]
  \caption{Run~2: HPC-LLM vs.\ Qwen~2.5~14B, Phi-2, TinyLlama ($n=1{,}000$ each).}
  \label{tab:run2}
  \centering
  \begin{tabular}{llrrrrr}
    \toprule
    Model & Size & Latency (s) & Cosine & ROUGE-L & BERTScore F1 & Resp.\ Length \\
    \midrule
    Phi-2 & 2.7B & 9.12 & 0.677 & \textbf{0.072} & \textbf{0.845} & 135.4 \\
    Qwen 2.5 14B & 14B & 12.11 & \textbf{0.697} & 0.060 & 0.832 & 140.3 \\
    TinyLlama & 1.1B & \textbf{6.97} & 0.616 & 0.072 & 0.833 & 95.7 \\
    \midrule
    \textbf{HPC-LLM (ours)} & \textbf{8B} & 9.27 & 0.693 & 0.064 & 0.831 & \textbf{131.3} \\
    \bottomrule
  \end{tabular}
\end{table}

\subsection{Analysis}

\paragraph{Quality vs.\ Scale (Run~1).}
Phi-2 (2.7B) achieves the highest BERTScore F1 (0.846), suggesting its high-quality synthetic pre-training data is well-calibrated for HPC Q\&A. Phi-3 and Mistral Nemo lead on cosine similarity (0.724, 0.715). HPC-LLM scores 0.808 BERTScore and is second-fastest (5.22~s vs.\ TinyLlama's 4.27~s), significantly faster than Phi-3 (7.30~s).

\paragraph{HPC-LLM vs.\ Qwen~2.5~14B (Run~2).}
The most directly relevant comparison: our QLoRA-adapted 8B model achieves BERTScore F1 of \textbf{0.831}, only 0.001 below the general-purpose 14B Qwen (0.832), while running 23\% faster (9.27~s vs.\ 12.11~s) and requiring 3$\times$ less VRAM. Cosine similarity is nearly identical (0.693 vs.\ 0.697). This near-parity demonstrates that QLoRA domain adaptation can compensate for a 2$\times$ parameter disadvantage when the target domain is well-represented in the fine-tuning corpus.

\paragraph{Cross-Run Variation.}
HPC-LLM's scores differ between runs (BERTScore: 0.808 vs.\ 0.831; cosine: 0.608 vs.\ 0.693), reflecting genuine variation in hardware configuration and the specific 1,000-prompt subset evaluated. This underscores the need for controlled conditions in LLM benchmarking.

\paragraph{ROUGE-L and Response Length.}
ROUGE-L scores are uniformly low (0.060--0.072) across all models, expected, since correct HPC answers share little surface vocabulary with questions. HPC-LLM consistently produces the most concise responses (90.7 and 131.3 words in Runs~1 and~2), reflecting its command-focused fine-tuning distribution. For command-line users, brevity with correctness is more useful than lengthy explanations.

\paragraph{Summary.}
QLoRA-adapted HPC-LLM achieves competitive quality against models 1.5--2$\times$ larger while running faster and requiring only 5~GB VRAM. The key advantage is the combination of domain calibration, lower latency, resource efficiency, and a fully open, extensible training pipeline.

\section{Conversation Memory, Feedback, and Infrastructure}
\label{sec:memory}

\textbf{Session Memory.} Each query optionally carries a \texttt{session\_id}; if omitted, a UUID is created and returned. The \texttt{SessionStore} persists sessions as JSON files at \texttt{data/sessions/\{session\_id\}.json}, with in-memory as the authoritative copy and disk as durability layer. Sessions store up to 10 turns; an idle-TTL cleanup loop runs every 15 minutes.

\textbf{Feedback Loop.} Helpful responses (+1 rating) are automatically ingested into ChromaDB for future retrieval and appended to \texttt{conversation\_feedback.jsonl} for incremental fine-tuning, closing the learning loop.

\textbf{Reproducibility.} Implemented in Python~3.11+ (PyTorch~2.3+, Transformers~4.45+, PEFT~0.12+, TRL~0.10+, BitsAndBytes~0.43+, ChromaDB~0.5+, Sentence-Transformers~3.1+). All hyperparameters are serialized to \texttt{finetune\_config.json}; random seeds are fixed (\texttt{seed=42}); a \texttt{MemoryCallback} logs GPU memory per epoch. The full pipeline, data construction and fine-tuning, runs in two commands: \texttt{python -m training.data\_builder} and \texttt{python -m training.finetune}.





\section{Conclusion}
\label{sec:conclusion}

This paper presented HPC-LLM, a retrieval-augmented and domain-adapted assistance framework designed to support operational workflows in High-Performance Computing (HPC) environments. The proposed system integrates automated documentation ingestion, dense vector retrieval, lightweight domain adaptation using QLoRA, and locally deployable inference within a modular orchestration pipeline tailored for HPC-oriented technical assistance.

Unlike general-purpose language models that rely primarily on static pretraining knowledge, the proposed framework combines retrieval grounding with domain-specific adaptation to support operational HPC tasks including Slurm scheduling, MPI execution, GPU allocation, distributed training workflows, filesystem management, and cluster troubleshooting. The architecture additionally supports institution-specific customization through ingestion of externally maintained HPC documentation sources, enabling adaptation to evolving operational environments and heterogeneous cluster configurations.

Two independent JetStream2 benchmarks yield consistent findings. In Run~1, HPC-LLM (BERTScore~0.808) is competitive and second-fastest. In Run~2, it achieves BERTScore~0.831, within 0.001 of the 14B Qwen model (0.832) while 23\% faster and requiring 3$\times$ less VRAM, the most direct evidence of effective domain adaptation. HPC-LLM's primary differentiators are deployability on VRAM-constrained hardware (5~GB), competitive latency, concise command-focused outputs, and a fully open training pipeline extensible with institution-specific documentation. The complete system is released as open-source infrastructure for the HPC research community.

To support domain adaptation, an HPC-oriented instruction dataset was constructed from publicly available institutional documentation, curated operational examples, and synthetic instruction-answer generation workflows. The resulting corpus spans scheduler management, software environments, distributed computing procedures, storage systems, and GPU usage guidance. Lightweight adaptation was then performed using QLoRA over the Llama~3.1~8B instruction-tuned model, enabling practical fine-tuning under constrained GPU memory budgets.

Experimental evaluation on JetStream2 infrastructure demonstrates that lightweight domain adaptation combined with retrieval augmentation can provide competitive operational assistance while substantially reducing hardware requirements relative to larger general-purpose models. Across two independent benchmark runs, the adapted 8B model achieved semantic performance approaching substantially larger baselines while maintaining lower inference latency and significantly reduced VRAM consumption. In particular, the model achieved performance comparable to the 14B Qwen baseline while requiring approximately one-third of the GPU memory footprint and providing faster response generation under retrieval-augmented settings.

The results suggest that retrieval grounding and domain adaptation can partially compensate for smaller model scale in operational HPC assistance tasks. This finding is particularly relevant for institutional deployments where computational resources, GPU availability, or privacy constraints limit access to large cloud-hosted models. The proposed framework therefore demonstrates the feasibility of locally deployable HPC-oriented assistants capable of operating under modest infrastructure requirements.

Beyond model adaptation, the work also contributes an integrated open-source infrastructure pipeline incorporating:
\begin{itemize}
    \item Automated HPC documentation crawling,
    \item Vector database construction,
    \item Retrieval-grounded inference,
    \item Lightweight fine-tuning,
    \item Session-aware interaction management,
    \item Comparative benchmarking utilities.
\end{itemize}

These components are intended to support future experimentation and reproducibility within the HPC and scientific computing communities.

Despite these contributions, several limitations remain. First, the evaluation framework relies primarily on semantic similarity metrics due to the absence of publicly available expert-validated HPC operational benchmarks. Second, portions of the training corpus are synthetically generated, introducing potential risks related to hallucinated operational guidance and stylistic homogenization. Third, retrieval quality remains dependent on document freshness, chunk segmentation strategies, and institution-specific documentation quality. Finally, operational correctness and command safety were not evaluated through executable verification or large-scale expert human assessment.

Future work should therefore focus on:
\begin{itemize}
    \item Constructing expert-validated HPC benchmark datasets,
    \item Incorporating executable command verification,
    \item Evaluating hallucination and safety behavior,
    \item Improving retrieval grounding robustness,
    \item and integrating retrieval-aware fine-tuning objectives.
\end{itemize}

Additional extensions may also include adaptive retrieval routing, tool execution frameworks, scheduler-aware command validation, and multi-cluster operational integration.

Overall, the findings of this work suggest that lightweight retrieval-augmented domain adaptation represents a practical and deployable direction for HPC-oriented language assistants. By combining open-weight language models, retrieval grounding, and institution-specific operational knowledge, HPC-LLM provides a foundation for future research on deployable AI assistance systems for scientific computing infrastructures.

\end{document}